\documentclass[11pt]{article}

\usepackage[style=ieee,backend=biber]{biblatex}
\usepackage{graphicx}
\usepackage{tabularx}
\usepackage{booktabs}
\usepackage[T1]{fontenc}
\usepackage{lmodern}
\usepackage[flushleft]{threeparttable}
\usepackage[hidelinks]{hyperref}
\hypersetup{hidelinks}
\usepackage{multirow}
\usepackage{longtable}
\usepackage[margin=3cm]{geometry}
\usepackage{amsmath}
\usepackage{amssymb}
\usepackage{authblk}
\usepackage{caption}
\usepackage{flafter}
\title{A Robust Camera-based Method for Breath Rate Measurement}
\author[1]{Alexey Protopopov}
\affil[1]{Joint Stock Research and Production Company Kryptonite \authorcr
E-mail: a.protopopov@kryptonite.ru}
\date{}
\begin{document}
    \captionsetup[figure]{labelformat={default},labelsep=period,name={Figure}}
    \captionsetup[table]{labelformat={default},labelsep=period,name={Table}}

    \maketitle

    \begin{abstract}
	Proliferation of cheap and accessible cameras makes it possible to measure a subject's breath rate from video footage alone. Recent works on this topic have proposed a variety of approaches for accurately measuring human breath rate, however they are either tested in near-ideal conditions, or produce results that are not sufficiently accurate. The present study proposes a more robust method to measure breath rate in humans with minimal hardware requirements using a combination of mathematical transforms with a relative deviation from the ground truth of less than 5\%. The method was tested on videos taken from 14 volunteers with a total duration of over 2 hours 30 minutes. The obtained results were compared to reference data and the average mean absolute error was found to be at 0.57 respirations per minute, which is noticeably better than the results from previous works. The breath rate measurement method proposed in the present article is more resistant to distortions caused by subject movement and thus allows one to remotely measure the subject's breath rate without any significant limitations on the subject's behavior.
    \end{abstract}


    \section{Introduction}\label{introduction}

Traditionally, most vital signs monitoring methods require some form of contact with the patients. However, thanks to the proliferation of cheap and accessible measurement devices, recent years have seen increased interest toward non-contact vital signs monitoring which possesses many inherent advantages over the traditional methods. For example, being able to monitor a patient’s life signs remotely reduces the probability of pathogen transmission between the patient and the medical personnel performing the measurement. Furthermore, such methods are more expedient since they do not require a sensor to be attached to the patient’s body. At the same time, non-contact vital signs monitoring presents a number of challenges which negatively affect the accuracy of such methods. In context of measuring a person’s breath rate (BR), the main challenge lies in filtering the artifacts caused by patient motion that is not related to breathing. Previous works on this subject have attempted to solve this problem and achieve an accurate BR measurement. For instance, in \autocite{vanEsch} the authors tried to develop a camera-based method of BR measurement for patients at hospitals and have achieved a mean absolute error (MAE) of 1.12~RPM. The method used cameras mounted above patient beds in intensive care units which collected records over the course of 24~hours, or until the patient was discharged. The authors did not specify how large the error value is in relation to the measured BR, but given that the average BR of a human at rest lies between 12 and 16~RPM we can assume that the relative deviation is close to 8\%. In another paper \autocite{Caroppo} the authors measured BR among elderly subjects that were asked to sit in front of a camera and look directly at it. They achieved a MAE of 1.80~RPM at a distance of 0.5~m between the camera and the subject’s face. As with the previous work the authors did not mention the relative error, but it can be assumed to be close to 13\%. It should also be noted that the result of these measurements was not a curve representing the change of BR over time, but rather a single value computed from a 3~minute long video. Finally, in \autocite{Wang} the authors achieved a MAE of 2.10~RPM in daylight conditions when testing their method on a phantom, rather than a person. The BR simulated by the phantom lied between 5 and 60~RPM, with the average being approximately 24~RPM, meaning that the relative error was close to 8\%. In general, all of the methods mentioned above demonstrated deviations from the ground truth that were above the 5\% threshold, despite being tested in conditions devoid of large-scale subject movement. Thus, in this paper we propose a more robust method of measuring a subject’s BR from a video recording made with a single web camera by calculating the visible motion of the subject’s chest region and filtering out interference.

\section{Methods}\label{methods}

We validated our findings using a dataset containing 14~video recordings gathered from 8~male and 6~female volunteers aged between 20 and 65 (informed consent was obtained from all participants) with a total duration of over 2~hours 30~minutes. The volunteers were selected solely based on their willingness to participate in the experiment, they were not selected based on their ethnicity. During the experiment the participants were asked to solve a simple puzzle on a computer. This was meant to distract the volunteers from their breathing and induce a natural respiration rate. These recordings were acquired using a Logitech 720p web camera positioned approximately 60~cm in front of the participant’s face.

Reference BR measurements were made using a pair of mechanical wearable breath sensors, with the upper sensor intended for thoracic breathing and the lower for diaphragmatic breathing. These measurements were synchronized with the video recordings to provide an accurate reference.

\subsection{Subject segmentation}\label{meth_subj_segm}

The aim of the present work is to develop a more robust method that would be able to accurately measure BR regardless of the subject’s movement in the video. Thus, it is not enough to select a region of interest (ROI) on one of the frames and apply it to the entire video. Instead, the ROI must be drawn dynamically on every frame and it must correspond to the subject’s chest region, since that is what breath motion is primarily associated with. For this purpose, Mediapipe multi-class selfie segmentation model was used to select the relevant regions of the subject’s body. This particular model segments the frame into a number of zones: background, hair, skin of the face, body skin, clothes and accessories. For the purpose of detecting breath motion the regions of the frame corresponding to clothes and body skin zones were merged to create a binary mask. Unfortunately, the chosen segmentation model turned out to be prone to what is best described as “zone jitter”, where the outline of the zones would rapidly change between frames. To counter that effect the method calculates the overlap between the binary masks of the past 10~frames, which is then applied to the frame. The pixels within the selected zones are left intact, while all other pixels are made black. 

It should be noted that of all the steps involved, segmentation is the most computation-intensive, being responsible to over 90\% of the processing time. However, the model used in the present work was tailored for mobile GPU operation and had to be run on a CPU. Using a different segmentation model to process the video on a GPU would significantly reduce running time.

\subsection{Motion calculation}\label{meth_motion_calc}

\begin{figure}[htbp]
    \centering
    \includegraphics[scale=0.7]{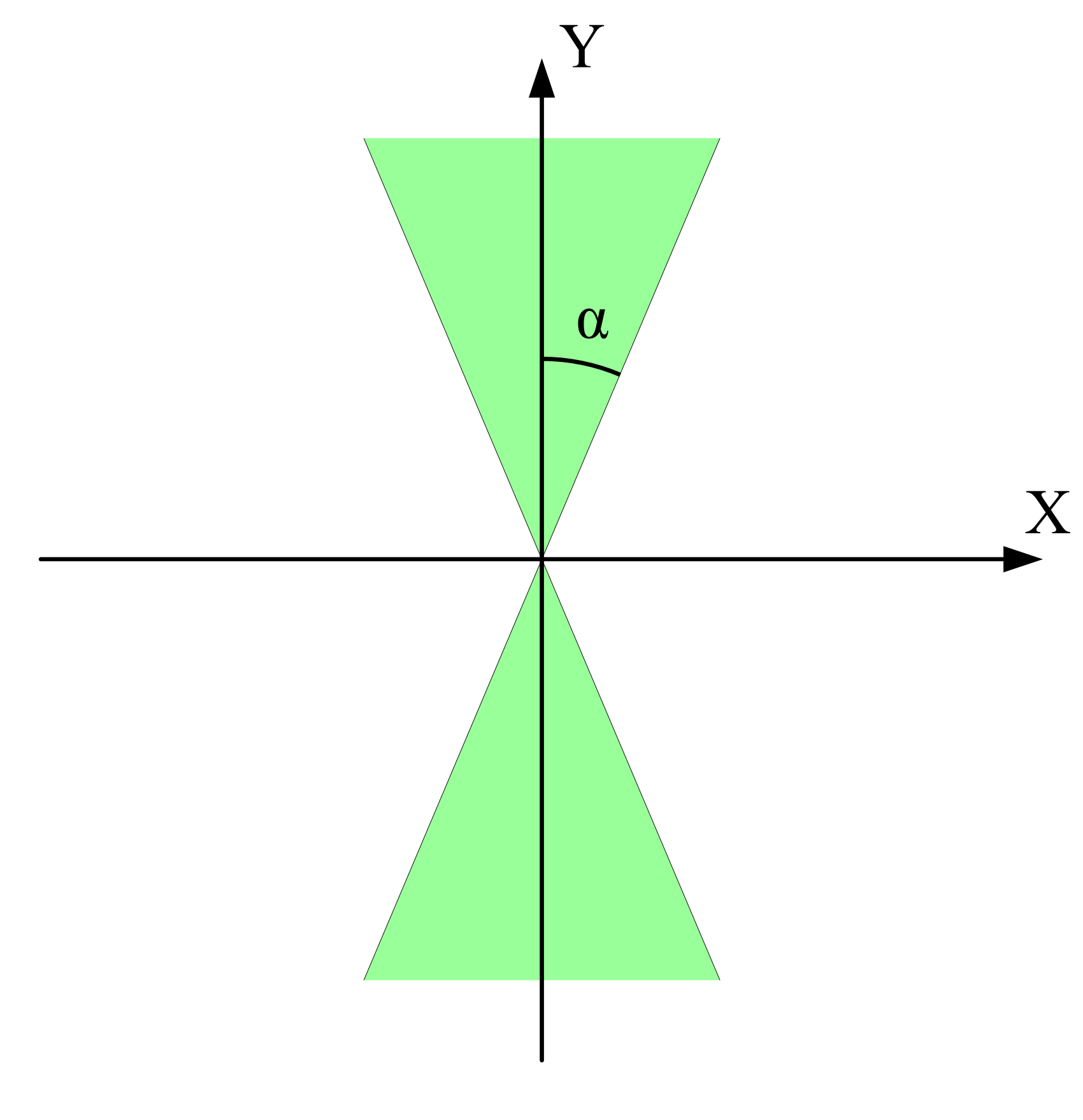}
    \caption{Direction filter for movement vectors. Vectors that deviate too much from the vertical axis were discarded.}
    \label{fig:1}
\end{figure}

After the mask was applied to the frame, the Farnebäck method \autocite{Farneback} was used to calculate the motion between frames, which produced a vector map, where each pixel was assigned a 2D vector corresponding to the movement of objects in its vicinity between the previous and current frame:

\begin{equation}
    \vec{v}_{i,j}=\begin{bmatrix}
    x_{i,j} \\
    y_{i,j}
    \end{bmatrix} 
    \label{eq:1}
\end{equation}

where \emph{x\textsubscript{i,j}} and \emph{y\textsubscript{i,j}} are the horizontal and vertical components of the vector, and \emph{i} and \emph{j} are the indices of the corresponding pixel along the horizontal and vertical axis. Given that the proposed method is expected to be applied to subjects sitting in front of a camera, it can be assumed that the motion related to their breath is predominantly vertical. Thus, interference from movement such as talking, leaning and waving one’s hands can be filtered out by applying a direction filter to the vector map, as shown on Figure~\ref{fig:1}.

Vectors that are not sufficiently vertical had their amplitude set to zero. Mathematically this means that the angle \emph{$\beta$} between the vector and the Y~axis must satisfy one of the following conditions:

\begin{equation}
	\begin{aligned}
	    &|\beta| < \alpha \\
	    &|\beta| > \pi - \alpha
	    \label{eq:2}
    	\end{aligned}
\end{equation}

where \emph{$\alpha$} is the threshold angle. In the present work this angle was set to 0.52~radians.

The remaining vectors were added together to produce the vector of the overall movement for this frame:

\begin{equation}
    \vec{V}=\sum_{i,j}\vec{v}_{i,j}
    \label{eq:3}
\end{equation}

Finally, we took this vector’s angle and used it to plot a time series, where each value corresponded to a frame of the video.

\subsection{Signal processing}\label{meth_signal_proc}

With the time series obtained, which will be referred to as the “signal”, we applied a series of filtering steps to remove unwanted noise. First, the signal was smoothed by averaging it across a sliding window 0.65~seconds wide, and a Butterworth lowpass filter was applied with the cutoff frequency of 0.496~Hz. Given that the period of the subject’s breath is expected to be longer than 1~second, this removed most of the noise without affecting the information related to breathing.

\begin{figure}[htbp]
    \centering
    \includegraphics[scale=0.7]{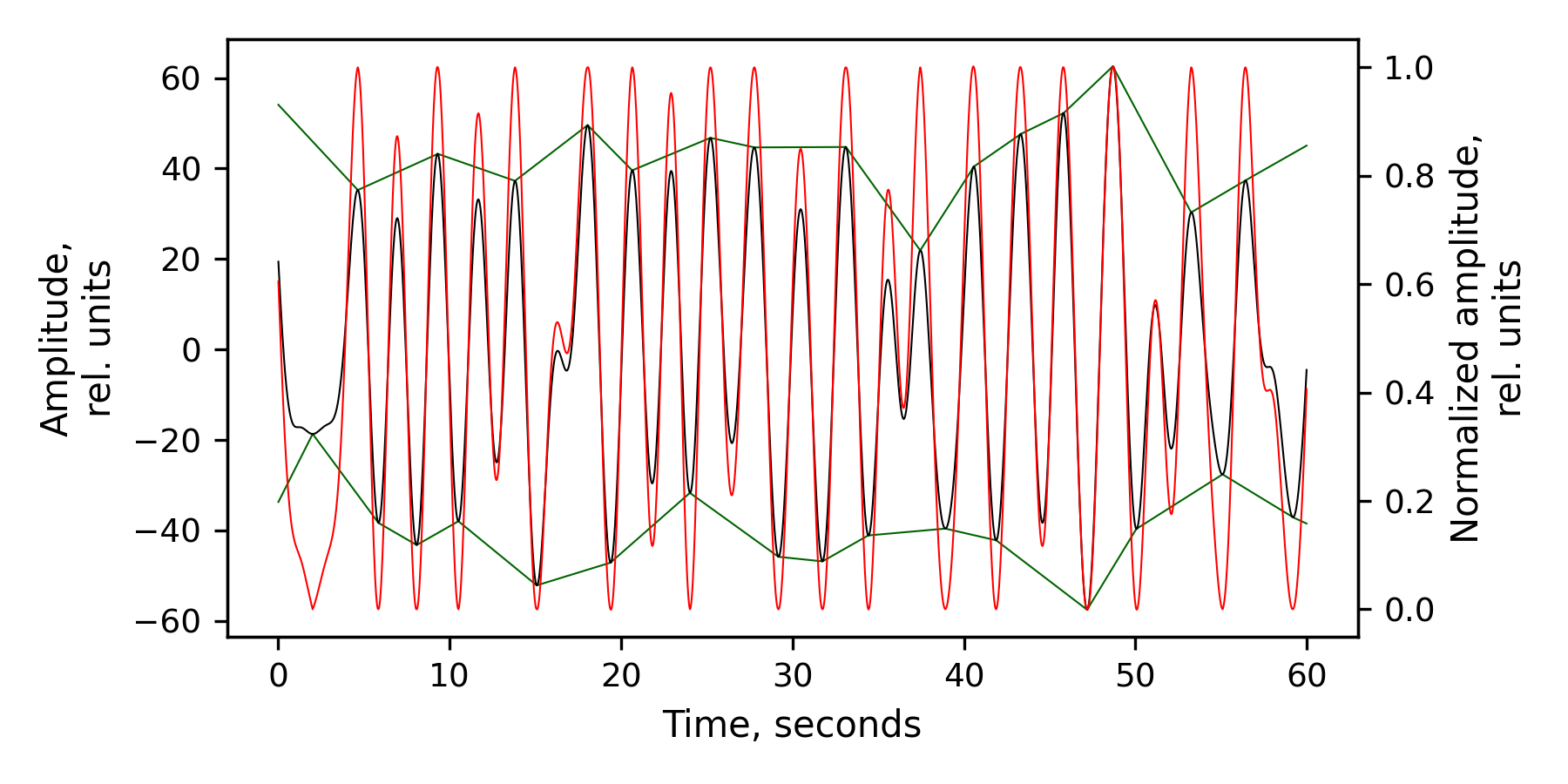}
    \caption{The result of signal normalization. The filtered signal is drawn in black, the polylines are drawn in green, the resulting normalized signal — in red.}
    \label{fig:2}
\end{figure}

The filtered signal was then normalized to an interval of [0;~1]. We achieved this by first locating the maxima and minima of the signal within a 6~seconds long sliding window and then using them as vertices of two polylines, one connecting the maxima, and the other connecting the minima as shown in Figure~\ref{fig:2}. 

The signal was then normalized using the following formula:

\begin{equation}
    \widehat{s}_{n}=\frac{s_{n} - p_{min,~i}}{p_{max,~i} - p_{min,~i}}
    \label{eq:4}
\end{equation}

where \emph{s\textsubscript{n}} is a sample of the filtered signal, \emph{$\widehat{s}$\textsubscript{n}} is a sample of the normalized signal, \emph{p\textsubscript{min, n}} and  \emph{p\textsubscript{max, n}} are the corresponding points of the polylines.

Finally, the normalized signal was fed into SciPy’s \emph{find\_peaks} function. The optimal parameters for this function were determined experimentally. Thus, minimum peak height was set to 0.496~relative units, peak prominence was 0.1848~relative units and the minimum distance between peaks was 1.5~seconds, which corresponded to BR of over 37~rpm — significantly higher than what was observed in the experiments. After the peaks were found, inter-peak intervals were calculated, averaged over a sliding window of 60~seconds and converted to BR in respirations per minute.

\section{Results and discussion}\label{results}

BRs calculated using the proposed method (video BR) were compared against BRs calculated using reference sensors (reference BR) and the results are presented in Table~\ref{tab:1}. The quality of the performed measurements is quantified with the following three values: mean absolute error (MAE) between video BR and reference BR, bias and root mean square deviation (RMSD). Overall, we achieved an average MAE of 0.57~rpm over all 14 experiments, which translates into an average relative error of less than 4\%. 

\begin{table}
    \caption{Experimental results.}
    \label{tab:1}
    \centering
    \begin{tabular}{||c|c|c|c|c|c||}
        \hline
        Experiment & Duration, min & Mean BR, rpm & MAE, rpm & Bias, rpm & RMSD, bpm \\
        \hline
        1  & 09:19 & 18.25 & 0.04 & 0.00 & 0.00 \\
        \hline
        2  & 08:53 & 17.32 & 0.40 & 0.18 & 0.37 \\
        \hline
        3  & 10:15 & 19.00 & 0.70 & 0.04 & 1.14 \\
        \hline
        4  & 08:22 & 17.91 & 0.77 & -0.20 & 1.23 \\
        \hline
        5  & 09:51 & 22.76 & 0.20 & 0.13 & 0.18 \\
        \hline
        6  & 10:38 & 19.01 & 0.72 & -0.23 & 1.34 \\
        \hline
        7  & 05:18 & 17.63 & 1.01 & -0.56 & 1.88 \\
        \hline
        8  & 10:43 & 15.70 & 0.58 & 0.32 & 0.86 \\
        \hline
        9  & 15:36 & 18.44 & 0.87 & 0.29 & 1.43 \\
        \hline
        10 & 15:46 & 21.65 & 0.95 & 0.16 & 1.55 \\
        \hline
        11 & 10:47 & 20.10 & 0.29 & 0.21 & 0.26 \\
        \hline
        12 & 12:00 & 22.33 & 0.27 & 0.18 & 0.22 \\
        \hline
        13 & 16:25 & 16.07 & 0.30 & 0.25 & 0.40 \\
        \hline
	14 & 15:31 & 18.15 & 0.75 & 0.00 & 1.04 \\
        \hline
    \end{tabular}
\end{table}

Figure~\ref{fig:3} shows a side-by-side comparison of a typical processed signal from the video recording and the lower reference sensor, before they are subjected to the peak finding function:

\begin{figure}[htbp]
    \centering
    \includegraphics[scale=0.6]{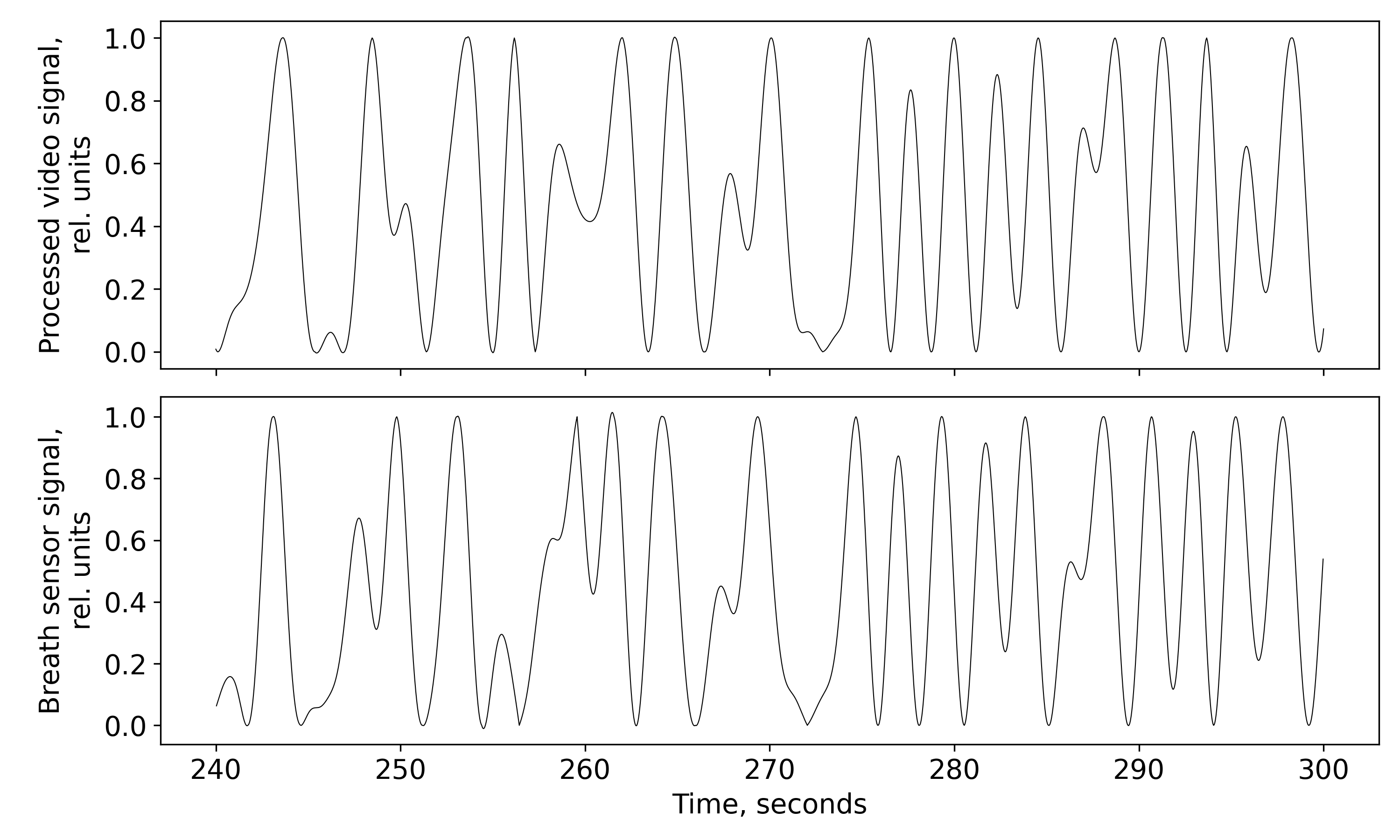}
    \caption{Comparison of a signal from the video to a reference signal.}
    \label{fig:3}
\end{figure}

Finally, Figure~\ref{fig:4} shows a comparison between the video BR and the reference BR:

\begin{figure}[htbp]
    \centering
    \includegraphics[scale=0.6]{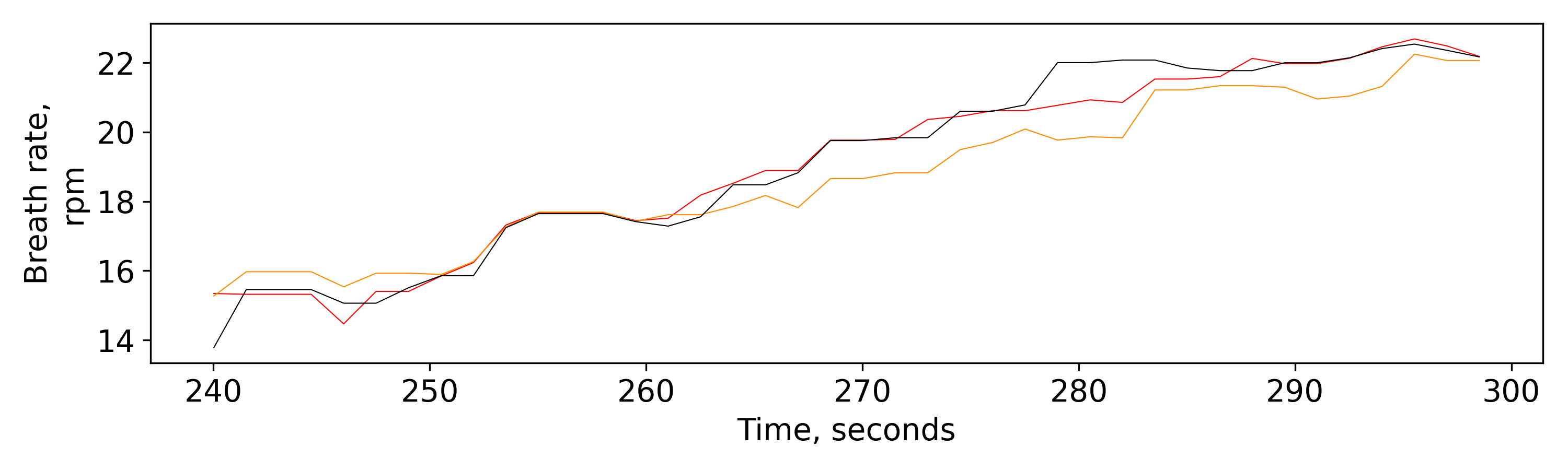}
    \caption{Comparison of the video BR to the reference BR. Black — BR acquired from the video, red — BR according to the upper breath sensor, orange — BR according to the lower breath sensor.}
    \label{fig:4}
\end{figure}

As can be seen in Table~\ref{tab:1}, BR calculated by the proposed method closely matches the one measured by the reference sensors. It is important to note here that the experiment imposed very few limitations on the subjects’ behavior: they were asked only to stay within the camera’s field of view while completing the puzzle and were not prohibited from talking or moving. Unsurprisingly, the recordings where the subjects talked during the experiment produced less accurate results than the ones where conversation was kept to a minimum.

A comparison between the proposed method and methods presented in existing works is given in Table~\ref{tab:2}, which shows that our method is superior to most existing methods in terms of MAE. In case of \autocite{Addison} however, the authors did not provide the value of MAE, instead opting for using a combination of bias and RMSD to gauge the quality of their approach. The deviation from the reference value in their work is somewhat lower than what we achieved, however it should be noted that the authors used a depth camera for their experiments, a device that is significantly more expensive than the web camera used in the present work, which brings the question of whether a slight increase in quality is worth such a large increase in price of equipment.

\begin{table}[h]
    \caption{Comparison to other works.}
    \label{tab:2}
    \centering
    \begin{tabular}{||c|c|c|c||}
        \hline
        Study & MAE, rpm & Bias, rpm & RMSD, bpm \\
        \hline
        Proposed method    & 0.57 & 0.10 & 0.85 \\
        \hline
        \autocite{vanEsch} & 1.12 & — & — \\
        \hline
        \autocite{Caroppo} & 1.80 & — & — \\
        \hline
        \autocite{Wang}    & 2.10 & — & — \\
        \hline
        \autocite{Mateu}   & 0.86 & — & — \\
        \hline
        \autocite{Addison} & — & -0.03 & 0.69 \\
        \hline
    \end{tabular}
\end{table}

Another thing to note is the computation speed. We have found that the time required to process a recording is approximately 10~times greater than the duration of the recording itself. Most of this time is spent on segmenting the frames, which is explained by the fact that the Mediapipe segmentation neural network used in the proposed method is intended for use on mobile devices, and since our computations were performed on a computer, it had to be run on a CPU, which severely limited its performance. Theoretically the selection of the algorithm used for segmentation should have a minimal effect on the quality of BR calculation. Therefore, substituting the current neural network for one that performs better should also improve the performance of the proposed method without loss of accuracy. Given sufficient computational power, the method could even be adapted to measuring BR nearly in real time, albeit with a 30~second delay caused by the length of the window used to average inter-peak intervals.

\section{Conclusion}\label{conclusion}

To summarize the present article, the proposed method has the following advantages:
\begin{enumerate}
    \item Robustness. The method displayed good performance despite the subjects moving and talking in the recordings;
    \item Low hardware requirements. Footage gathered by virtually any web camera would be suitable for the proposed method;
    \item Potentially adaptable to real time measurements. The main factor limiting the speed of calculations is the segmentation neural network, which may be replaced in favor of one more suitable for the situation.
\end{enumerate}

At the same time, the method is not without flaws: 
\begin{enumerate}
    \item The time required to process a video is 10 times greater than the length of the said video;
    \item The method has a limit on the minimum length of the recordings, which cannot be shorter than the length of the window used to average inter-peak intervals, which is set to 60~seconds in the present paper.
\end{enumerate}

In its present state, the proposed method can already be used in a number of applications, the most obvious being for quickly measuring the patient’s vital signs during a doctor appointment, especially combined with existing methods of camera-based heart rate measurements \autocite{Protopopov}. Such measurements could also be performed without the presence of medical personnel by the patient himself using a smartphone, which could allow some appointments to be conducted remotely. Furthermore, the proposed method can be adapted for measuring BR of drivers to help in detection of drowsiness and loss of concentration.

\section{Acknowledgements}\label{acknowledgements}

I would like to thank my supervisor Vasiliy Dolmatov for his guidance and support, which greatly helped my research.

\clearpage
\printbibliography

\end{document}